\documentclass{llncs}
\usepackage{amsmath}

\usepackage{makeidx}  
\usepackage{graphicx}
\usepackage{url}
\usepackage{paralist}
\usepackage[ruled,vlined,noresetcount]{algorithm2e}
\usepackage{bbding}
\usepackage{soul, color}
\sethlcolor{green}
\usepackage{xspace}
\usepackage{algorithmic}
\usepackage{verbatim}

\RequirePackage{xspace}

\newcommand{\ourCL}{{\sc CoSMo}}

\DeclareMathAlphabet\mathbfcal{OMS}{cmsy}{b}{n}

\usepackage{tabularx}

\begin{document}
\title{\ourCL: A constructor specification language for Abstract Wikipedia's content selection process}
\author{Kutz Arrieta\inst{1}, Pablo R. Fillottrani\inst{2,3}, C. Maria Keet\inst{4}}
\institute{
Independent scholar, USA, \email{kutzaki@gmail.com}
\and Departamento de Ciencias e Ingenier\'ia de la Computaci\'on, 
Universidad Nacional del Sur, 
Bah\'ia Blanca, Argentina, \email{prf@cs.uns.edu.ar}
\and Comisi\'on de Investigaciones Cient\'{\i}ficas, Provincia de Buenos Aires, Argentina
\and
Department of Computer Science, University of Cape Town, South Africa,
            \email{mkeet@cs.uct.ac.za}
     }       

\maketitle               

\begin{abstract}
Representing snippets of information abstractly is a task that needs to be performed for various purposes, such as database view specification and the first stage in the natural language generation pipeline for generative AI from structured input, i.e., the content selection stage to determine what needs to be verbalised. For the Abstract Wikipedia project, requirements analysis revealed that such an abstract representation requires multilingual modelling, content selection covering declarative content and functions, and both classes and instances.
There is no modelling language that meets either of the three features, let alone a combination. 
Following a rigorous language design process inclusive of broad stakeholder consultation, we created \ourCL, a novel {\sc Co}ntent {\sc S}election {\sc Mo}deling language that meets these and other requirements so that it may be useful both in Abstract Wikipedia as well as other contexts. We describe the design process, rationale and choices, the specification, and preliminary evaluation of the language.
\end{abstract}

\setcounter{footnote}{0}

\section{Introduction}
\label{sec:intro}

Databases technologies have been evolving from relational database management system (RDBMS) silos to also comprise RDF triple stores that can be queried on the Web more easily as Linked Data that may be distributed and accessible through a SPARQL endpoint, and other graph-based approaches, such as 
knowledge graphs 
\cite{Hogan20}. Conceptual modelling and accessing such data may be expected to evolve with it, such as with virtual knowledge graphs and ontology-based data access (e.g., \cite{FK20,Xiao19}), where the model serves the dual purpose of abstract representation of the data store's content and as starting point for query construction. One such publicly editable data store is Wikidata, which, among other things, is intended to be used in the new Abstract Wikipedia project \cite{Vrandecic20,Vrandecic21}. Abstract Wikipedia aims to be a `next generation' truly multilingual Wikipedia, where articles are generated from data stored in Wikidata (and functions from Wikifunctions) in at least all supported languages of the Wikipedia ecosystem. This requires essentially an open, publicly editable multilingual natural language generation system for verbalising structured data, information, and knowledge that is selected from Wikidata and Wikifunctions.

Such a system is currently under development\footnote{\url{https://meta.wikimedia.org/wiki/Abstract_Wikipedia}} and initial related research is being conducted, such as verbalising all RDF statements in a given domain into basic English \cite{Amaral22} and {\em ad hoc} opaque content selection but with a considerable emphasis on multiple languages with Grammatical Framework \cite{Ranta23}. What is supposed to be positioned there, is what the community refers to as ``constructors'' and the ``abstract representation'' (see fn. 1, \cite{Vrandecic20}), for which no proposal exist yet, only an exploration of what it may, or may not, involve\footnote{\url{https://meta.wikimedia.org/wiki/Abstract_Wikipedia/Wikidata_Abstract_Representation}; version of 29 Nov 2022}. The selection of content from Wikidata, and possible processing with functions from Wikifunctions, is expected to be done by end users, in snippets or small modules (the basic `constructors') at a time, where the constructors can be put together to automatically generate larger pieces of text to make up an article on a topic in the desired language, and declaring such constructors ideally would be done in the user's preferred natural language. This content selection step is suggestive of being alike creating a database view, but with a few extras, notably calculating something, such as the age based on date of birth. That is, it needs at least a `view creation' syntax for RDF, alike SQL's {\tt CREATE VIEW}, but in such a way that it is implementation-independent, since the Wikidata database may be converted to JSON or moved into an RDBMS for performance at a later stage and one would not want to have to update all user-defined constructors because of it. Alternatively, such a prospective constructor language can be seen as one aimed at information modelling.

\begin{figure}[h]
\centering
\includegraphics[width=1.0\textwidth]{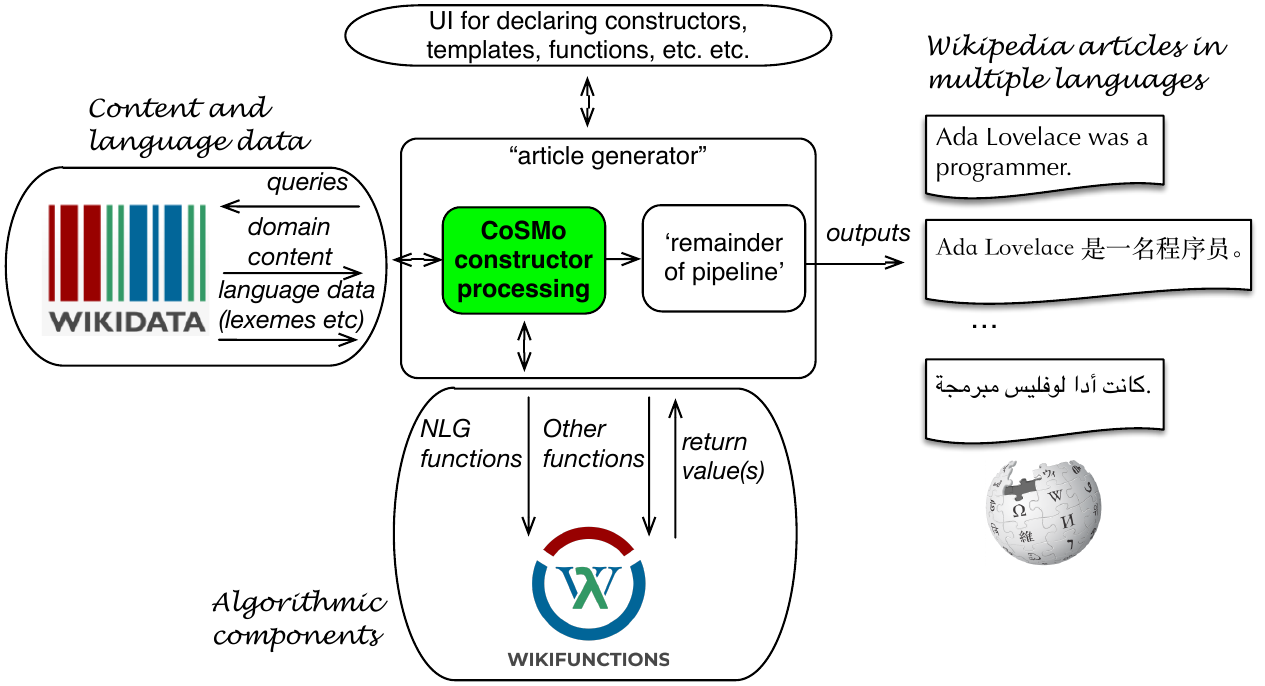}
\caption{Sketch where \ourCL\ fits in a high-level architecture for Abstract Wikipedia.}
\label{fig:overview}
\end{figure}

There is no modelling language that meets these requirements. 
In this paper we propose a new modelling language for the constructors, i.e., the content selection specification step, which can be used with both Abstract Wikipedia, 
positioned as illustrated in Fig.\ref{fig:overview}, 
and other content selection settings in at least, the natural language generation pipelines for verbalising structured data, information, and knowledge, and other tasks where users need to select a small fragment of a model. The key features and novelties compared to other conceptual data modelling and view specification languages are:
\begin{itemize}
\item Multilinguality is embedded in the modelling language; 
\item A fully modular approach, where modules (as so-called `constructors') are first-class citizens in the language;
\item Declarations can be made at both class and instance level in the same language;
\item Declarations can contain both static content and functions to compute content that can be derived from the static content.
\end{itemize}
We followed a rigorous language design process that was first introduced in \cite{FK20fois,FK21}, and as such this is its first industry-initiated use case. We describe the design process, rationale and choices made, the specification of the language---called \ourCL, {\sc Co}ntent {\sc S}election {\sc Mo}deling language---and a preliminary evaluation. The evaluation showed that it meets these and other requirements so that it may be useful both in Abstract Wikipedia as well as other contexts. 

The remainder of the paper is structured as follows. Section~\ref{sec:relwork} provides background on natural language generation, Abstract Wikipedia, and the notion of modelling language design. Section~\ref{sec:langdesign} describes the process followed, design principles chosen, and the specification of \ourCL. Section~\ref{sec:eval} reports on the evaluation of \ourCL. We discuss and compare it to related modelling languages and graphical query formulation solutions in Section~\ref{sec:disc} and conclude in Section~\ref{sec:concl}. 

\section{Background}
\label{sec:relwork}

The related work consists of three converging themes: features and limitations of current conceptual data modelling languages, procedures for the design of modelling languages, and natural language generation in general and for Abstract Wikipedia in particular. 

\begin{figure}[h]
\centering
\includegraphics[width=1.0\textwidth]{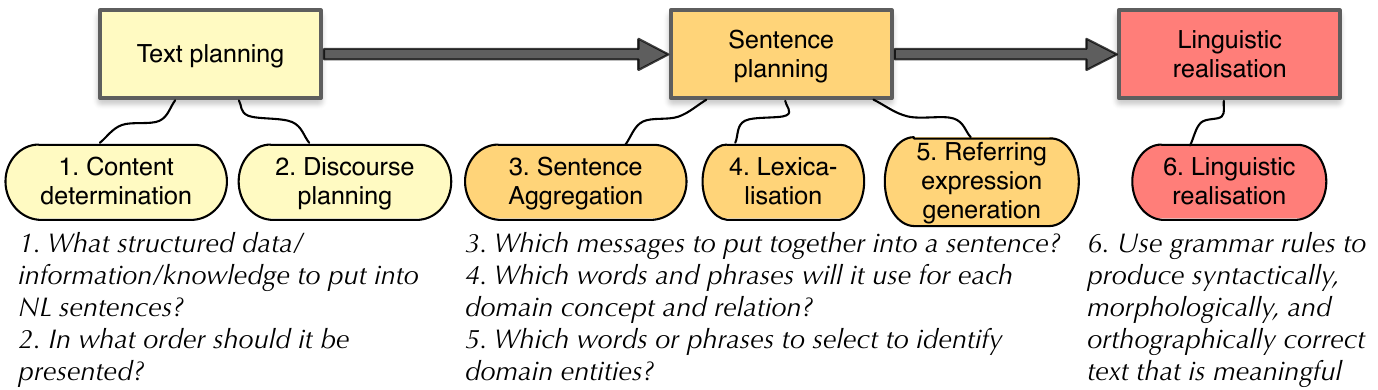}
\caption{The reference NLG pipeline of Reiter and Dale. For basic EER/UML/ORM conceptual data model verbalisation, it may instantiate as: 1: the model; 2: e.g., classes first, then their attributes, then their relationships; 3: as in 2; 4 and 5: the vocabulary used in the model, possibly sentence aggregation; 6: none or minor fixing with camel case and spaces.}
\label{fig:pipeline}
\end{figure}

There are many ways to generate natural language from structured input. The structured input is typically either data, such as from spreadsheets and databases, or structured information, such as verbalising conceptual data models, or structured structure knowledge, which includes verbalising ontologies. The desired output can be as varied, from instructions to educational questions, to generating text summaries of soccer matches. What they all have in common is a pipeline approach. The `reference pipeline' of Reiter and Dale \cite{Reiter97} is a useful framework since all systems step through it, even when consciously deciding to skip a task or merge two tasks,
or loop between sub-tasks. 
This pipeline is shown in Fig.~\ref{fig:pipeline}. Steps 1 and 2 are on the side of input processing, whereas steps 3-6 are about getting desirable text as output. It is also possible to add additional sub-pipelines, such as a pipeline specific to step 6, realisation, only (see \cite{MK23} for a recent overview).

Within the context of Abstract Wikipedia, the input is to be extracted from Wikidata, the computations are expected to be done on Wikifunctions, and the output is to be a Wikipedia article \cite{Vrandecic21}. Wikidata is an RDF triple store with over 14 billion triples. Selecting content from it to create an article is, in general idea, akin to declaring a view over a small part of the RDF triple store, but then with the additional option to declare functions, such as computing a person's age from their date of birth that is stored in Wikidata. Currently, there is no way to do so other than writing individual SPARQL queries at its SPARQL endpoint. It lacks a way of declaring what Abstract Wikipedia calls the `constructors'. The two proof-of-concept tools, Ninai/Udiron \cite{Morshed23} and the one with Scribunto, use a temporary mechanism, where the former merges data structures from across the pipeline whilst it resolves query answering through the SPARQL endpoint for the Wikidata content and lexeme data, and the latter has dummy constructors to focus on the realisation component\footnote{Ninai/Udiron by Mahir Morshed: \url{https://gitlab.com/mahir256/ninai/-/tree/main/ninai/constructors}; Scribunto-based one by Ariel Gutman: \url{https://meta.wikimedia.org/wiki/Abstract_Wikipedia/Template_Language_for_Wikifunctions/Scribunto-based_implementation}.\label{fn:ninai}}. Also Grammatical Framework aims to be useful in the NLG pipeline for Abstract Wikipedia. That system focuses on multilingual surface realisation and also, to date, has no solution for the content selection step \cite{Ranta23}, other than a `verbalise all triples' and a sketch of a suggestion to use a Python overlay for declaring queries in one natural language, naming them, and then integrating them in a Python template (section 8 in \cite{Ranta23}). Thus, the critical steps of systematic, robust, multilingual, and end-user accessible content selection and ordering is missing from the overall Abstract Wikipedia pipeline to date, and thus also where something needs to be chosen from the design options.

The following example illustrates the process assuming the prospective wiki ecosystem. It is not an optimal scenario, but intended to illustrate in an intuitive way certain possible design decision that can be made along the way in the pipeline.

\begin{example}\label{ex:DA}
Let us consider a snippet of the Wikidata knowledge graph, with the tiny fraction of the 14 billion triples listed in short-hand notation, and natural language annotation added afterward to indicate the labels of the Q and P items, with Q42 being Douglas Adams:

\begin{small}
\begin{verbatim}
....
<Q42, P31,Q5>  // Douglas Adams (Q42) instance of (P31) human (Q5)
<Q42, P106, Q214197> // has occupation (P106) playwright (Q214197)
<Q42, P106,Q18844224> // occupation science fiction writer
<Q18844224, P136, Q24925> // genre (P136) science fiction (Q24925)
<Q18844224, P2521, "autorka sci-fi (Czech)"> // female form of label
...
\end{verbatim}
\end{small}
As can be seen, Q items can refer to instances (e.g., Q42) and to classes (e.g., Q5 and Q18844224) and the property labels may be a reified version (e.g., P106) of a verb phrase (e.g., read `has occupation' or `works as').

The task of the {\em content selection} step is twofold: 1) to select an `interesting' part of the graph that is relevant to the user and 2) optionally to add any functions to it, such as a ranking of results or computing the age of an object and so on. This sort-of view over the graph is our (first) constructor. This can be done for a particular instance, such as retrieving information about Q42/Douglas Adams specifically that is deemed relevant by the constructor author, or retrieving information about all science fiction writers/Q18844224. It is precisely this data structure and language for it that does not exist yet that we will introduce in the next section. For this illustration, let us write something that looks intuitive enough to follow this small example (although having several shortcomings):
\begin{small}
\begin{verbatim}
Constructor-DA(
    Object(Q42){            // Douglas Adams
          property(P31)     // instance of (P31) 
          property(P106)    // occupation (P106) 
          property(P136)    // genre (P136)
       })
\end{verbatim}
\end{small}
Assuming a suitable parsing of this constructor, it retrieves that Douglas Adams is a human, all occupations that he had (Wikidata lists 8 occupations) and all genres linked to that (which includes science fiction). 

Second, the pipeline moves on to {\em discourse planning}, or: in which order would one want to write or say this in natural language? Several orderings are possible, making it a 1:n relation between what's happening in step 1 and step 2 
of the pipeline. 
It is also possible to group step 2 together with later steps in the pipeline, such as whether it will be rendered in one sentence or in more than one. In this example, we will keep it separate to illustrate it. For instance, we may prefer an ordering as {\tt RendererConstructor1}, {\tt 2} or {\tt 3} and merge it with whether it will be verbalised in one or in more than one sentence:
\begin{small}
\begin{verbatim}
RendererConstructor1(
 Object(Q42), property(P31). Object(Q42),property(P106),property(P136))
    
RendererConstructor2(
 property(P136),property(P31),Object(Q42),property(P106))    
    
RendererConstructor3(
 property(P106),Object(Q42). property(P136). Object(Q42),property(P31).)      
\end{verbatim}
\end{small}
It is also here or afterward that some further verbalisation properties may be added, such as whether it should be in present or past tense and whether a referring expression should be generated, like not having twice {\tt Object(Q42)} in {\tt RendererConstructor1}, but, e.g., a {\tt Z12345(Object(Q42))} function stored on Wikifunctions so that the Q-item label in the second sentence is replaced by ``he'' or ``his'' (in this example). It might then look like, e.g.: 
\begin{small}
\begin{verbatim}
RendererConstructor1re(
 Object(Q42), property(P31). Z12345(Object(Q42)), property(P106),
     property(P136))
\end{verbatim}
\end{small}
Finally, such a constructor has to relate to templates or a grammar-based realiser for the different natural languages for the last {\em linguistic realisation} step, which also may be a 1:n relation. Two hypothetical (and simplified) templates, where the ``[ ]'' denote slots to be filled by content from the constructor and the content it obtained from Wikidata through that, just for {\tt RendererConstructor1}, might be:
\begin{small}
\begin{verbatim}
template1forRC1(
 [Q42] is an [P31] [rangeofP31]. [Q42] had as [P106]: [rangeofP106] 
    in the [P136] of [rangeofP106].)
    
template2forRC1(
 [Q42] was a [rangeofP31]. His [P106] include [rangeofP106] with as 
    [P136] [rangeofP106].)
\end{verbatim}
\end{small}
It then still may need some final processing in the linearization step, depending on the slot's content and the language, such as  language-motivated phonological operations (e.g., `a' vs `an') and capitalization, to then finally produce that sentence. The {\tt template1forRC1} would generate ``Douglas Adams is an instance of human. Douglas Adams had as occupations: playwright, screenwriter, novelist, children's writer, science fiction writer, comedian, writer, musician in the genre of science fiction.''. Arguably, this is not the most exciting or easily readable sentence, with this walkthrough only illustrating a sampling of the decisions to be taken and what that then looks like. In praxis, the NLG systems are generally more sophisticated than this illustration. \hfill $\diamondsuit$
\end{example}

\begin{figure}[h]
\centering
\includegraphics[width=1.0\textwidth]{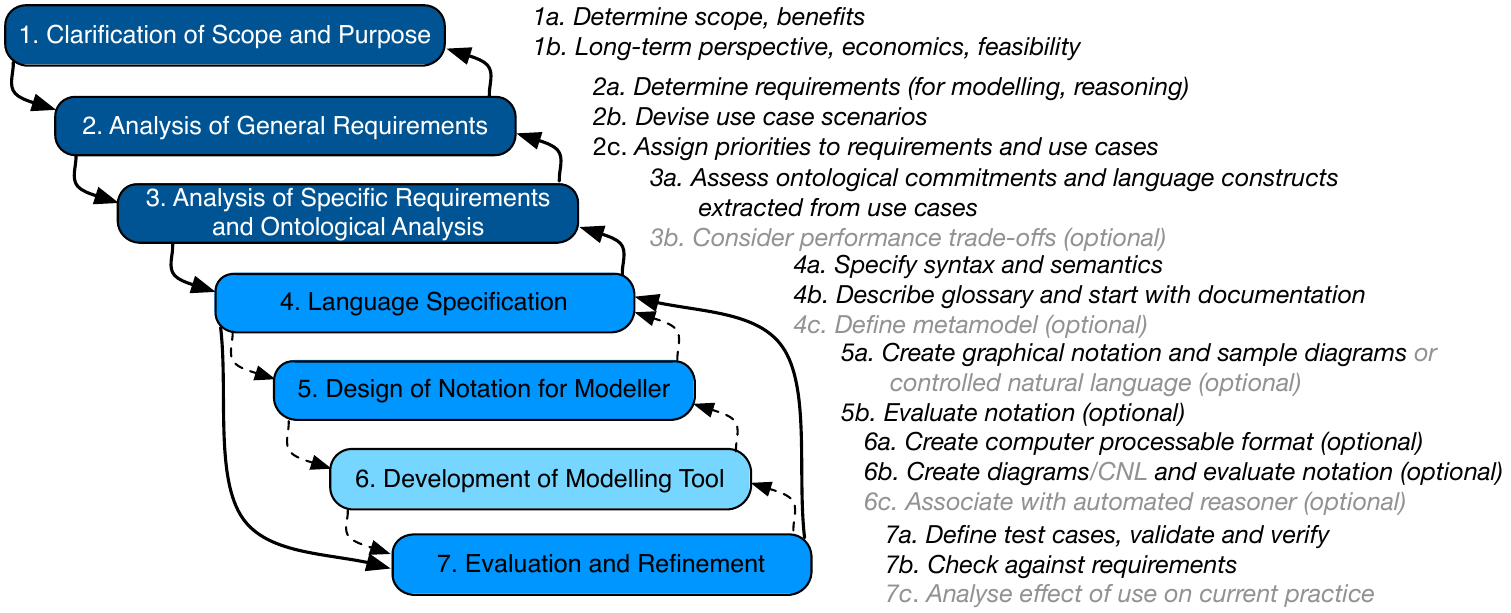}
\caption{CDML development process of \cite{FK21}. The figure is adapted to the Abstract Wikipedia case as follows: dark blue: wide consultative process; medium blue: actual language creation by the authors; light blue: planned; grey text: task not used because deemed not needed.}
\label{fig:langdes}
\end{figure}

It is that language for the constructors in step 1 that we aim to design, not a language for realisation nor a template language. This entails a notion of language design.  
Language design procedures have received little attention in the conceptual modelling 
and knowledge representation 
communities, despite that a number of languages exist with multiple notational variants and numerous extensions thereof. This is to some extent explainable by having moved the goalposts of what the modelling language should be able to do, but it may also be the case that there are so many variants because the first versions of those languages did not go through a rigorous design process. Either way, at least a design procedure is needed. In our prior work on design processes, we modified Frank's waterfall model \cite{Frank13} to an iterative one that is  informed by ontology \cite{FK20fois}, with one version specific to ontology language design \cite{FK20fois} and one adapted to conceptual data model design \cite{FK21}. This is merged and summarised in Fig.~\ref{fig:langdes}. The various steps were illustrated in \cite{FK20fois,FK21}, but the procedure has not been tested with an industry use case. 

The next example illustrates the process briefly, continuing from Example 1.

\begin{example}
If the {\tt Constructor-DA} were to be a real example as a use case scenario (step 2b of the procedure in Fig.~\ref{fig:langdes}), one should infer possible language requirements from it (step 2a). The constructor exemplifies a so-called frame-based or perspectivist view: there is one key object, Q42 in this case, and it has whatever properties we selected. 
The constructor  
also could have been, say,
\begin{small}
\begin{verbatim}
Constructor-Writing(
          property(P31)
          property(P106)
          property(P136)
          P106:range={Q18844224}           
          P31:domain={Q42}           
       )
\end{verbatim}
\end{small}
which might nudge a discourse and template editor to more flexibility. It still can handle the previous renderers from Example 1, but just as well invites other sentence constructions where the properties take centre stage. That is, the {\tt Constructor-Writing} is not from the object viewpoint in the same way as {\tt Constructor-DA} is. Reuse is also easier, since it is clear in the last two lines that a selection is made and other Qitems are then only a minor tweak. 

Continuing with the illustration of requirements extraction: both constructor samples show only P and Q items from Wikidata, so one might be in the (mistaken) belief that Wikifunctions functions are not needed in the content selection constructors, and add to a list of requirements the (correct) assumptions of multilinguality in content and need for label processing of items. \hfill $\diamondsuit$
\end{example}

Finally, a review of features of conceptual data modelling languages could fill a paper. Instead, we shall compare them against the distinctive features of the modelling language that we will propose in the next section, and therefore defer the related work and comparison to the discussion in Section~\ref{sec:disc}. We include a comparison with visual SPARQL builders there as well.

\section{Language design}
\label{sec:langdesign}

The original expectation for the constructor language is that it definitely will take care of step 1, content selection, of the NLG pipeline, and possibly step 2 and later as well for the part of linking up modules/constructors in some ordering so that the generated text won't be all randomly positioned in the article. We demarcate the language to be intended for step 1 of the NLG pipeline; whether modules in later steps of the NLG pipeline devise a separate data structure for discourse planning and other language information or inject new content in the same constructor by means of a strict language extension, is a separate question that is not within the scope of this paper. We first describe the procedure followed and then the specification.

\subsection{The processes of steps 1-3 to elicit requirements}

The methodological approach followed is overlayed over the diagram in Fig.~\ref{fig:langdes}. First, those steps out of scope were set aside, notably the CNL for the modelling language, since it is superseded by the prospective NLG pipeline, and the automated reasoner with performance trade-offs. Either may still be of interest to consider at a later stage if the language will be used for other purposes.

For the first three steps in the development process, several stakeholder meetings were held online on Google Meet in September and October 2022. An initial design document was drafted by one member (Kutz Arrieta) to which other members contributed and which was discussed and extended in the meetings and over email. These meetings comprised both theory-based approaches and example-based approaches. This was subsequently reworked into two documents, one for public consumption and discussion on the Abstract Wikipedia metawiki\footnote{\url{https://meta.wikimedia.org/wiki/Special:MyLanguage/Abstract_Wikipedia/Wikidata_Abstract_Representation}, d.d. 29 Nov 2022.} and one internal document with `leftovers' to be discussed yet. Around the same time one of the authors (Maria Keet) created a list of possible requirements emanating from the examples list and additional examples were devised. In parallel, two `exploration through implementation' approaches were pursued, which are the  aforementioned Ninai/Udiron system and the Scribunto-based proof-of-concept. Finally, further investigation into roles was pursued \cite{Keet23} and an open meeting was held on 21 Feb 2023 to explain the features list in order to clarify terminology across specialists from the different disciplines and obtain more feedback on preferences\footnote{recording available at: \url{https://meta.wikimedia.org/wiki/File:Abstract_Wikipedia_Natural_language_generation_working_group_-_2023_February.webm}.}. 
The latter meeting was deemed the `cut off' point to proceed to select the required and desired parameters, goals, and paradigms, and proceed to step 4.

\subsection{Step 4: Language Specification}

Broad {\em a priori} parameters exist for the Abstract Wikipedia project, which can be recast as goals and high-level requirements that the `abstract representation language' has to meet. Notably, they include adherence to Wikimedia Foundation founding principles and guidelines, an open source licence, availability and editability in all languages of the Wikimedia projects, and with, ultimately, the local Wikipedia communities being in control of the content\footnote{For details, see \url{https://meta.wikimedia.org/wiki/Abstract_Wikipedia/Requirements}.}. This has a partial overlap with the original design principles of the Web Ontology Language OWL on paper \cite{Horrocks03}, but the emphasis on multilinguality and community, in the sense of anyone (not just the modeller or logician communities), is stronger in the sense of a mandatory requirement rather than an optional goal to aspire. It has, to the best of our knowledge, no overlap with any broad design parameters for either of the conceptual data modelling languages or any logic. 

\subsubsection{Design principles}\label{sec:design}
Language feature requirements can be separated into principles that have to be met, i.e., on the `what' that has to be in the language, and the `how', or requirements on how that will be incorporated in a language. For instance, one may require roles (relationship components, association ends, 
argument positions), 
but there may be different ways to incorporate them, such as unnamed with an automatic counter for identifier, requiring a human-readable name, or selection from a pre-defined list of roles. A same list of principles may thus still result in vastly different representation languages. Nonetheless, they help setting tighter bounds than the broad parameters. Based on the Abstract Wikipedia meetings, presentations, discussions, proof-of-concept tools, feature comparison document, and additional examples, the principles chosen are as follows, in arbitrary order.
\begin{itemize}
\item Roles of relationships, not the Entity-Attribute-Value style. This aligns well with existing information modelling languages as well as the language and linguistic view on constructors where roles are used heavily for predicates and verbs, such as in VerbNet \cite{Palmer17} and FrameNet \cite{Baker98}.
\item Separation of the information modelling from the natural language rendering aspects, i.e., to aim for not having linguistic markers (in the broad sense) in the constructors.
\item For a first version of the language at least: any noise in the data is expected to be corrected in the source rather than building verification mechanisms into the modelling language. This is a deviation compared to the aim of constraints in conceptual data modelling languages (CDMLs), which reflects its position in the pipeline, it being after adding data to the database rather than before and that modifying the RDF triple store is open to anyone within the Wikibase parameters.
\item Vocabulary creation in constructors is prohibited; rather, Wikidata lexicographic content or Q or P item labels are to be used and if it is not in Wikidata yet, then the user must add it first before adding it to a constructor. This enables its multilinguality and prevents universe of discourse semantic mismatches and costly mapping assertions.
\item The language must permit both declarative aspects as well as functions, where a constructor may have either or both sort of content. This bears a resemblance to UML class diagram's attributes and methods, but then realised differently (e.g., function execution is assumed to happen on Wikifunctions). 
\item The representation should be the least perspectivist possible, so that fewer constructors need to be specified that are more versatile in their use in the NLG pipeline.
\item To facilitate the multilinguality of the language yet provide stability to the language specification, there must be a 1:1 mapping on reserved strings in the language specification to Wikidata items or lexemes. The key benefit of the 1:1 mapping is that then the language is {\em de facto} multilingual, because it can fetch the label in one's desired language from Wikidata, making it the first  multilingual CDML. This ensures that if an item is deleted in Wikidata, then the language will not be broken as knock-on effect. 
\item Constructors can be reused and combined in other constructors and they can be defined for either particular instances or at the type-level where then all the Wikidata instances that satisfy it are retrieved (and thus multiple articles or paragraphs or sentences can be created at once or a value can be set for a particular rendering such that only one item with its properties will satisfy it).
\item Constructors can be refined or generalised, in that one can take an existing constructor that a contributor declared and add or remove features, respectively. 
\end{itemize}
This list of decisions implicitly also communicates key topics that have been investigated and discussed several times during the course of requirements gathering stage. The last one on constructor size, reuse, and instance versus type level caused quite some confusion in the meetings, since many wanted type level constructors but typically started with instance-level examples and declarations. After much deliberation as to the constructor size, it was concluded that there is no unambiguous way to demarcate it. 
The size of a constructor will depend on
the amount of generalisation and compositionality for reusability of constructors.  
This is related to the issue of whether to add a reserved wrapper element like, say, `article': what may end up as an article in one language 
may be only a paragraph of an article in another language. Different stances might be taken to improve readability or tolerate NLG systems that cannot yet compute sentence aggregation and referring expressions. In addition, when it comes to realisation in a specific language, further down in the pipeline, the same constructor might be realized in one word (in highly agglutinative languages), a compound, a phrase, a long sentence or a sequence of sentences. This will also depend on the maturity of the realizer. In version 1
of the NLG component of Abstract Wikipedia, 
we could see a more telegraphic style realizer, which  facilitates generation while preserving content.

What was not extensively discussed during the meetings, is whether the language should be ontology-driven, in part due to aversion to the idea by some and it seems to be a general tendency within the wiki projects as it being too prescriptivist. An ontology-driven way can take several forms, however. For instance, it may be alike OntoUML where each class is stereotyped with an entity from the foundational ontology UFO \cite{Guizzardi18}, at the level of the language itself by embedding ontological commitments into it \cite{FK20fois}, both, or in a yet different way. Other ways could be, e.g., an optional alignment to a foundational ontology category by the user to any element of any of the widely-used foundational ontologies (with mappings at the back-end), or run as a background process either where the ontology operates at the level of Wikidata rather than the stage of the constructor creation or afterward on constructor checking as a behind-the-scenes automated verification step.
Either way, ontological commitments cannot be avoided for the language itself. The design principle and requirement of roles mean a so-called {\em positionalist} stance on the nature of relations rather than ``standard view'' or ``anti-positionalist'' \cite{Fine00}. The separation of the subject domain semantics from the language features also has implications for how one assumes the world to be.

Other features are either more contentious or less certain about possible ramifications or level of desirability. They can be designed as strict extensions to a base language, in a way in the opposite direction as the design of OWL and OWL 2 where strict fragments were specified of the expressive ideal. Here, a lean language that is simple to use is preferred, since community adoption is aimed for, but if a more experienced constructor author wants to add a more precise specification, they should be able to. This could be offered in an `advanced interface' cf. a `standard' interface for the language that meets the goals listed above.
Sample extensions include: 
\begin{itemize}
\item Mandatory participation of an element. This imposes the requirement that both that data must be available in Wikidata and there must be NLG algorithms in the target language to verbalise at least that part of the constructor.
\item Union of Q or P items. It could be argued that duplication should be cleaned up from Wikidata and therefore this would not be needed, or it could be achieved by merging two otherwise the same constructors. The former may not be easy to effectuate and for the latter, the union operator amounts to a simple shortcut syntactic sugar to simplify constructor creation.
\end{itemize}

Negation did not appear in any of the examples and thus made it in neither the principal nor the extended list. It may be an unexplored omission; as it is certainly not a conscious `no'. 

\subsubsection{The language: specification and notation for modellers}

Practically, this worked out in iterations, specifically: 1) specification (step 4), 2) commence design of visual notation (step 5), 3) discussion of examples, 4) extension of specification, followed by 5) extension of notation and creation of two variant notations to compare, 6) discussion of extensions and notations, 7) update of textual notation,  8) proceed to formalisation, 9) one more iteration over the graphical and textual notations in parallel with the formalisation (i.e., iterating between steps 4 and 5), 10) evaluation with a few use cases to experiment with notation (steps 5 + 7), and finalise all. What follows in the remainder of the section is the final specification and final modeller notation proposed.

\begin{table}[t]
	\centering
	\caption{\ourCL~basic elements. 
	The longform notation displays the English labels for readability, where the respective Wikidata items are listed in Table~\ref{tab:wikimap}. The formal syntax is given in appendix \ref{app:longsyntax}.}		\label{tab:elements}	
		\begin{tabular}{||>{\centering\arraybackslash}c|>{\ttfamily\scriptsize}p{2.8cm}|>{\scriptsize}p{6.5cm}||}
\hline
\multicolumn{1}{||c|}{{\bf Element Icon}} & \multicolumn{1}{p{3cm}|}{{\bf  Longform representation}} & \multicolumn{1}{c||}{{\bf Usage}} \\
		\hline \hline
\includegraphics{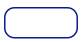}& Object(Q) \textrm{or} ObjectType(Q) &  Typically a Q (possibly a P) object (i.e., instance) or object type (i.e., class) \\ \hline  	
\includegraphics{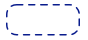}& Function(Z($\ldots$)) & $n-ary$ function to compute something that is not an NLG-related function (e.g., age), Z ‘object’. The function arguments have the representation given in the following row. \\ \hline  
\includegraphics{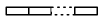}& Z(Q1,$\ldots$,Qn)
 &  Notation for $n$-ary function arguments; here, the $n-1$ first element correspond to the arguments for the function call while the last compartment connects to the function icon (indicated by the dashed connection here, as usual in ORM)
  \\ \hline 

\includegraphics{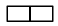}& Property(P(Role1, Role2))
 &  Typically a P (possibly Q) item, i.e., a property that relates (at least) two things. 
  \\ \hline 
\includegraphics{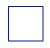}&TypeConstructor:T() & Constructor (at the type level), must be used in conjunction with several (at least one) definitions. T is a local variable   \\ \hline  
\includegraphics{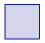}& InstanceConstructor:O() & Constructor (at the instance level), it must be used in conjunction with several (at least one) definitions. O is a local variable   \\ \hline
		\end{tabular}
\end{table}

We now present the syntax for \ourCL. This presentation is structured in three parts: the basic elements, the connectors between these elements, and the adornments that extend the expressivity of both elements and connectors. The basic elements are listed in Table~\ref{tab:elements}, where the first two columns contain the graphical and the longform textual version of \ourCL. A short version of the syntax is described in Appendix~\ref{app:shortnotation}. The graphical icons were selected as in ORM2 whenever is possible. The exceptions are the two constructor icons, one for types and one for instances which are absent in ORM2. The textual representation of each element in the table only shows its English label,
for presentation succinctness.
In addition to the Wikidata and Wikifunctions items, we can observe there is only the possibility to define new 
constructor
types and objects. The third column provides some details on how these elements can be used. The $n$-ary constructor is just a composition constructor of the $n-1$ function arguments plus the $n-1$-ary function connected to the last compartment.

Table \ref{tab:connectors} shows possible connectors between two of the basic elements in the previous table.  Again, graphical icons from ORM2 and UML were selected when possible. The only new icon is the trident icon  linking a new type or instance to its definition. This definition must be given in basis 
of 
a relationship participation, and is done by means of local variables. In order to foster simplicity of the language, we define the scope of these variables to be local to the current model. Global variables like in RDF would increase the reusability of the models, but this is not a current requirement for the language.  

\begin{table}[h]
	\centering
	\caption{Constructor language connector  vocabulary of \ourCL. The longform notation displays the English labels for readability, where the respective Wikidata items are listed in Table~\ref{tab:wikimap}.} 	\label{tab:connectors}				
		\begin{tabular}{||>{\centering\arraybackslash}p{2cm}|>{\ttfamily\scriptsize}p{3.5cm}|>{\scriptsize}p{5.7cm}||}
\hline
\multicolumn{1}{||p{2cm}|}{{\bf Connector Icon}} & \multicolumn{1}{c|}{{\bf Longform representation}} & \multicolumn{1}{c||}{{\bf Usage}} \\
		\hline \hline
\includegraphics{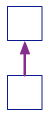}& SubConstructorOf(T1, T2) &  Subtype constructor between local variables  \\ \hline  	
\includegraphics{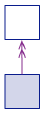}& InstanceOf(O, T) &  Instance  constructor between local variables\\ \hline  
\includegraphics{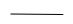}&  role:Object(Q)  {\rm or} role:ObjectType(Q) {\rm or}
Function(Z) & Connector/link to connect a Q to a P, or a  Z to a list of arguments  \\ \hline

\includegraphics[width=0.1\textwidth]{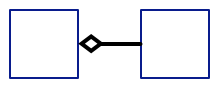} &  PartOf(T1,T2) & Connector/link for a container type {\tt T1} to a contained type {\tt T2}, both represented as a local variables.   \\ \hline
\includegraphics[width=0.1\textwidth]{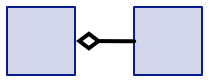} &  PartOf(O1,O2) & Connector/link for a container object {\tt O1} to a contained object {\tt O2}, both represented as a local variables.   \\ \hline

\includegraphics[width=1.2cm]{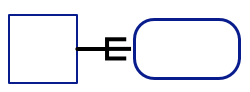} & TypeConstructor:T({\sl definition}) & Connector/link to define a new type  as a participation in a relationship; {\sl definition} contains
relevant predicate and role declaration \\ \hline
\includegraphics[width=1.2cm]{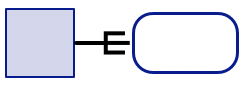} & 
InstanceConstructor:O({\sl definition}) & Connector/link to define a new object as a participation in a relationship; {\sl definition} contains relevant predicate, role and instantiation declarations. \\ \hline
		\end{tabular}
\end{table}

Finally, Table~\ref{tab:adornments} presents the constraint representations of \ourCL: value constraints, role names, join relationships, role mandatory participations and instantiations. Actually, role names are not constraints but just
syntactic sugar that later may be useful in further steps in the process.

\begin{table}[h]
	\centering
	\caption{Constructor language adornment  vocabulary of \ourCL.}		\label{tab:adornments}			
		\begin{tabular}{||>{\centering\arraybackslash}p{2cm}|>{\ttfamily\scriptsize}p{3cm}|>{\scriptsize}p{6cm}||}
\hline
\multicolumn{1}{||p{2cm}|}{{\bf Adornment Icon}} & \multicolumn{1}{p{3.5cm}|}{{\bf Longform representation}} & \multicolumn{1}{c||}{{\bf Usage}} \\
		\hline \hline
\includegraphics{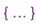}& \textrm{definition}\{ValueConstraint\} &  Value constraint for a type, an object or a function definition. Constraints are expressed as in ORM2
  \\ \hline  
  \includegraphics{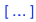} & Role[Name]:Object(Q) \textrm{or} Role[Name]:ObjectType(Q) & Defines the role as in the declaration, assigning a name to the participation. The definition of object or object type as in the first row of table \ref{tab:elements}.\\ \hline 
\includegraphics{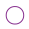}& Join(Q1, Q2) \textrm{or} Join(P1, P2) &  Merge/join between two objects or two relations of the same arity, within the definition of a type or an object.\\ \hline 
\includegraphics{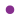}& IsMandatory(Role) &  Mandatory participation of the representation of the element for the constructor to be allowed to be realised, within the definition of a type or an object.\\ \hline 
\includegraphics{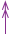} & ObjectType(Q1)=\{Q2\} & Instantiation constraint for the characterisation of an instance, relating one object QItem to another QItem representing a type, within the definition of a type or an object. \\ \hline
 		\end{tabular}
\end{table}

In order to specify the semantics, we formalize the meaning of the textual representation  in first order logic. Our language is formed with one constant $q, p, z$ for each P, Q, Z items referenced in Wikipedia and Wikifunctions; one unary predicate for each Q item, and each object variable, object type variable, and role variable; one binary predicate for each P item; and one $n$-ary function for each $n$-ary Z item. 
 We also include the special predicates ${\tt QItem()}$, ${\tt PItem()}$, ${\tt ZItem()}$, ${\tt Has()}$, 
 ${\tt PPartOf}()$ and ${\tt Contains}()$. The first three are unary predicates for each type of Wikidata items. 
${\tt Has}()$ is a set of predicates to represent the function application to a given set of arguments.  
${\tt PPartOf}()$  is a binary predicate formalising the proper parthood relation (assuming ground mereology \cite{Varzi04}), which is transitive, irreflexive, and asymmetric. 
${\tt Contains}()$ is quaternary  predicate relating an object in the domain with the predicate name and predicate instances it reifies. Remember that the only possibility of introducing a new element is by defining it as a particular predicate instantiation. The semantic mapping is shown in Table \ref{tab:semantics}. In a given model definition, we can have object and type constructor, subconstructor and instance definitions. Within the  type constructor we can have at least one predicate declaration, two or more role declarations, and zero or more function declarations. Role declaration may include object and object type declaration, with optional role names and value constraints.    The object constructor has the same structure as the type constructor, with the addition of at least one instantiation declaration.

\begin{table}[t]
	\centering
	\caption{Semantics of \ourCL\ given as first order logic formulae.	}	\label{tab:semantics}	
		\begin{tabular}{||>{\ttfamily\scriptsize\centering\arraybackslash}c|>{\scriptsize}p{6.5cm}||}
\hline
\multicolumn{1}{||c|}{{\bf Longform representation}} & \multicolumn{1}{c|}{{\bf FOL representation}} \\
		\hline \hline
 Object(Q) & ${\tt QItem}(q)$ \\
 ObjectType(Q) & $ {\tt QItem}(q)$\\
 ObjectType(Q1)=Q2 & ${\tt QItem}(q_1)\wedge {\tt QItem}(q_2) \wedge Q_1(q_2)$\\
 Property(P(Role1,Role2))& ${\tt PItem}(p)\wedge \forall x,y (P(x,y) \rightarrow Role1(x)\wedge Role2(y))$ \\
    Role:ObjectType(Q) & $\forall x ( Rol(x) \rightarrow Q(x))$ \\ 
     Function(Z(Q1,$\ldots$,Q2))   & ${\tt Zitem}(z)\wedge  {\tt Has}(z,q_1,\ldots,q_n,Z(o))$\\
    SubConstructorOf(Q1,Q2) & $\forall x(Q_1(x)\rightarrow Q_2(x))$\\
    InstanceOf(O,T) & $\forall x(O(x)\rightarrow T(x)$\\
    PartOf(T1,T2) & $\forall x(T1(x)\rightarrow \exists y(T2(y)\wedge {\tt PPartOf}(x,y)))$\\    
    TypeConstructor:T({\rm definition on $p$}) & $\forall x(T(x) \rightarrow (\exists y_1,y_2{\tt Contains}(x,p,y_1,y_2)\wedge \text{definition on tuple $p(y_1,y_2)$} )$\\ 
    InstanceConstructor:O({\rm definition on $p$}) &$\forall x O(x)\rightarrow (\exists y_1,y_2 {\tt Contains}(x,p,y_1,y_2) \wedge \text{definition on tuple $p(y_1,y_2)$} )    $\\
    Join(Q1,Q2) & $\forall x (Q(x)\leftrightarrow (Q1(x)\lor Q2(x)))$\\
    Join(P1,P2) & $\forall x,y (P(x,y)\leftrightarrow (P1(x,y)\lor P2(x,y)))$\\    
    \hline
		\end{tabular}
\end{table}

 A \ourCL~interpretation  is a tuple ${\cal I}=\langle \Delta, \cdot^{\cal I}, ob, name\rangle$. We assume a countable infinite domain $\Delta$, which includes all P, Q, Z items referred in Wikipedia and Wikifunctions, which we call $P$, $Q$ and $Z$. Thus, $P\cup Q\cup Z \subseteq \Delta$. 
 There is also in $\cal I$ a unique object assigned to each tuple by the injective function $ob:P\times Q\times Q \rightarrow \Delta$ assigning a unique domain element which objectifies  each predicate tuple, and a 
The interpretation function $\cdot^{\cal I}$ is defined as usual in predicate logic, with the additional following axioms: 
\begin{align*}
QItem(q)^{\cal I}  &\equiv  q^{\cal I}\in Q^{\cal I}\\
PItem(p)^{\cal I}   &\equiv  p^{\cal I}\in P^{\cal I}\\
ZItem(z)^{\cal I} &\equiv z^{\cal I}\in Z^{\cal I}\\
Q_1(q_2)^{\cal I} &\equiv  q_2^{\cal I}  \in Q_1^{\cal I}\\
P(x,y)^{\cal I} &\equiv (x^{\cal I}, y^{\cal I}) \in P^{\cal I}\\
Role(x)^{\cal I} &\equiv x^{\cal I} \in Role^{\cal I}\\
Has(f,x,...,x,r)^{\cal I} &\equiv  r^{\cal I}=f^{\cal I}(x^{\cal I},\ldots,x^{\cal I})\\
Contains(x,p,y_1,y_2)^{\cal I} &\equiv ob(p^{\cal I},y_1^{\cal I},y_2^{\cal I})=x^{\cal I}
\end{align*}
Then, all interpretations that satisfy a given content selection model definition are models for that definition. These models may be useful for the following steps in the NLG pipeline.

\section{Evaluation}
\label{sec:eval}

The main aim of the evaluation is to test the expressiveness of the language: can it do what it should do? To examine this, we need to check \ourCL~against the requirements and design decisions and revisit the examples that were used to devise the set of language requirements to determine whether they can be represented in the new modelling language. We shall first illustrate an `easy' case to show how the language may be put to use, then revisit the requirements, and finally examine a selection of the example constructors from the requirements gathering stage. 

\subsection{Example constructors}

\begin{figure}[h]
\centering
\includegraphics[width=0.8\textwidth]{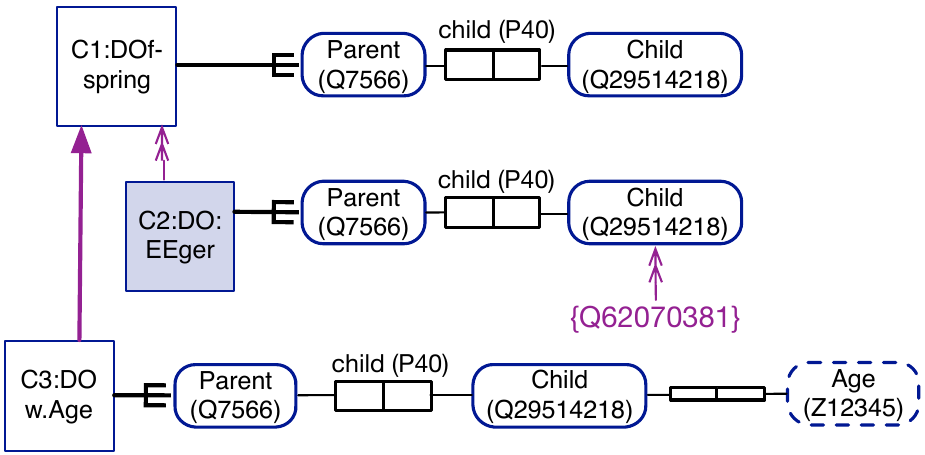}
\caption{Two type-level constructors ({\sf C1} and {\sf C3}) and an instance-level constructor ({\sf C2}). Names are placeholders, which are expected to be assigned an identifier in its own namespace and with at least one label.}
\label{fig:EEEx}
\end{figure}

Let us commence illustrating the first example from the Abstract Representation discussion document on the metawiki, about Edith Eger, and for illustrative purpose, add a few variations. First, Edith Eger is a child of two parents, for which we can create a type-level constructor ({\sf C1}) and instantiate it where Edith Eger (Q62070381 in Wikidata) specifically is selected, in {\sf C2}, which is shown in the top-half of Fig.~\ref{fig:EEEx}. The naming and numbering of the constructors are {\em ad hoc} in this example, standing in for the expected decision that constructors will get their own namespace in the Wiki ecosystem. Also the number for the function is made up, since Wikifunctions is in beta testing at the time of writing. Subtyping of a constructor is illustrated with {\sf C3} being a subconstructor of {\sf C1}, since it has an additional {\sf Age} function. The instantiation and subtyping may also be drawn in different figures.

\noindent The longform serialisation of the diagrams, or straight written in text format, can be as follows:

\begin{verbatim}
TypeConstructor:C1(
   Property(P40(r1,r2)),     
   r1:ObjectType(Q7566), 
   r2:ObjectType(Q29514218))

InstanceOf(C2, C1)

InstanceConstructor:C2(
   Property(P40(r1,r2)),     
   r1:ObjectType(Q7566), 
   r2:ObjectType(Q29514218), 
   ObjectType(Q29514218)={Q62070381})

SubConstructorOf(C3, C1)
   
TypeConstructor:C3(
   Property(P40(r1,r2)),     
   r1:ObjectType(Q7566), 
   r2:ObjectType(Q29514218),
   Function(Z12345(Q29514218)))
\end{verbatim}

\noindent Alternatively, one may wish to use shorthand notation, so that the constructors are specified thus:
\begin{verbatim}
TC:C1(
   P40(r1,r2), r1:Q7566, r2:Q29514218)

IC:C2(
   P40(r1,r2), r1:Q7566, r2:Q29514218, Q29514218={Q62070381})
   
TC:C3(
    P40(r1,r2), r1:Q7566, r2:Q29514218, Q29514218.Z12345)
\end{verbatim}
Neither constructor has their roles named yet, with only local, within-constructor, numbering of roles (assuming an auto-counter). It is possible to name both roles, which, at this stage of modelling, are `ontological' (non-linguistic) roles (see \cite{Keet23} for an analysis). A possible improvement of {\sf C1} may be where {\sf r2} is named with Q239526 ({\sf offspring}/{\sf descendencia}/{\sf avkomma}/etc. labels for all entered languages), leaving {\sf r1}  as-is:
\begin{verbatim}
 TC:C1( 
   P40(r1,r2), r1:Q7566, r2[Q239526]:Q29514218)   
\end{verbatim}

Finally, by the grammar specification of \ourCL, one can see that the constructor may be larger than the single fact types shown here. For reusability of constructors, specifically for Abstract Wikipedia, it would be good practice to keep the constructors `small', meaning where its contents would be verbalised in one short paragraph at most.

\subsection{Meeting the requirements}

We now revisit the requirements and design principles to determine if, and if so where, they have been met, roughly in the order as listed in Section~\ref{sec:design}.

First, roles are present and first-class citizens in \ourCL. As is customary practice in conceptual data modelling, they may be named with subject domain-specific names (a Q item from Wikidata), but need not \cite{Keet23}. There are no linguistic markers in the \ourCL~specification, nor content verification checking, and by design, one cannot create new vocabulary {\em in} the constructor. The constructor itself needs an identifier and at least one label to ensure its multilinguality as well, which requires an adaptation of Wikidata, in line with the ongoing deliberations on constructors\footnote{\url{https://meta.wikimedia.org/wiki/Abstract_Wikipedia/Updates/2023-03-15}}. The mapping of the language elements to Wiki's Q/P items to ensure multilinguality is listed in Table~\ref{tab:wikimap}. 
At present, it is being deliberated in the community what the best implementation strategy would be: to either reuse the indicated tentative Q and P items and add the remainder of items and their respective labels, or to add new Q and P items specifically for \ourCL\ elements. For this reason, as well as possible Wikidata-independence to increase \ourCL's potential for wider application use and for languages currently not yet supported by Wikidata, \ourCL\ elements have their own identifiers that are eventually mapped to Wikidata items. 

\begin{table}[h]
	\centering
	\caption{Constructor language vocabulary of \ourCL, with identifiers and labels in three languages. The Q and P items are assumed to become at least {\em de facto} immutable (or `locked' from edits) in the same way as other key elements in Wikidata and Wikifunctions. TBA: Q item yet to be added to Wikidata; text in italics: item label yet to be added to Wikidata.}			
		\begin{tabular}{|p{1.4cm}|p{2.1cm}|p{2.6cm}|p{3cm}|p{2.5cm}|}
\hline
{\bf \ourCL\ ID} & {\bf \textit{Tentative} Q/P item in Wikidata} & {\bf Feature (English label)} & {\bf Example of label in Spanish} & {\bf Example of label in Basque} \\
		\hline \hline
CSM001 & Q35120 ?? & Object& Entidad & Izaki\\ \hline  	
CSM002& (TBA) & {\em Object type} & {\em Tipo de entidad} & {\em izaki mota} \\ \hline  	
CSM003& Q18616576 & Property& Propiedad & propietatea\\ \hline  	
CSM004&Q117747915 ?? & Role& {\em Rol} & {\em Rol}\\ \hline  	
CSM005&Q11348 & Function& Funci\'on & Funtzio \\ \hline  	
CSM006&Q1049476 ?? & {\em Instance Constructor} & {\em Constructor de instancia} & {\em instantzia eraikitzaile} \\ \hline  	
CSM007& (TBA) & {\em Type constructor} & {\em Constructor de tipo} & {\em tipo eraikitzaile} \\ \hline  	
CSM008& (TBA) & {\em Subconstructor of}& {\em Subconstructor de} & {\em honako berreraikitzaile}\\ \hline  	
CSM009&Property:P31 & InstanceOf& Instancia de & honako hau da \\ \hline  	
CSM010&Property:P527 & Part of& {\em Parte de} & Osatuta \\ \hline  	
CSM011&Q17853087 & Join& Uni\'on & {\em bildura} \\ \hline  	
CSM012&Q29557567 & {\em IsMandatory} & {\em Es obligatorio} & {\em nahitaezko}\\ \hline  	  
CSM013& Q82799 & Name& Nombre & Izen \\ \hline  	
CSM014& Q42750658 & Value constraint& Restricci\'on de valor & {\em balio murrizte}  \\ \hline  	 	 		 		 		 		 
		\end{tabular}
	\label{tab:wikimap}
\end{table}

Also, \ourCL~has the requisite elements for both declarative components, being any P and Q item, and for functions, by incorporating the Z identifiers from Wikifunctions. It does this in such a way that it need not be perspectivist for the declarative components at least: they are all elements on par in an arbitrary, re-orderable sequence. The functions are of something, and  do need to relate to something, which can be denoted as if on par in the long notation, but in the shorthand notation with the dot-notation, it is tied to a Q item. Further functionality includes the specialisation/generalisation and instantiation of constructors, each indicated with a different type of arrow in the diagrammatic notation and an additional shaded box. Instead of overloading line notation, we have added an additional line type to connect constructors to an element of its content.\footnote{Drawing a box around the constructor's content is more appropriate conceptually and logically. We tried, but decided against it because it cluttered the view too much, hampering reading and understanding the diagrams. One is, of course, free to design a different diagrammatic notation.}

Finally, the two `extras' were added as well. The mandatory participation of an element is indicated with the purple dot. The union of items, if not done through combining two constructors, can be done in a constructor with the open circle/{\tt Join} element. 

\subsection{Revisiting other use case examples}

A precise reconstruction of all sample constructors used in the requirements elicitation is not possible, with the key reasons due to certain design choices, such as the ``no new ad hoc vocabulary'', whereas a few sample constructors allowed, and have, that. Another reason is that it has long been unclear what ``the abstract representation language'' is supposed to accomplish, which had revealed widely different assumptions during the requirements elicitation phase. One such key difference is whether it was to reside close to the linguistic realisation stage of the pipeline and thus needed to contain natural language and grammar cues, or early on in the pipeline and therewith natural language independent content selection, or cater for all stages. The use case examples reflect that diversity. Therefore, in revisiting them, we will show what parts can be catered for with \ourCL~and what has to be addressed elsewhere in the prospective  Abstract Wikipedia system, as well as an indication how that might be done (which, however, depends on the NLG system chosen).

\begin{figure}[h]
\centering
\includegraphics[width = 0.85\textwidth]{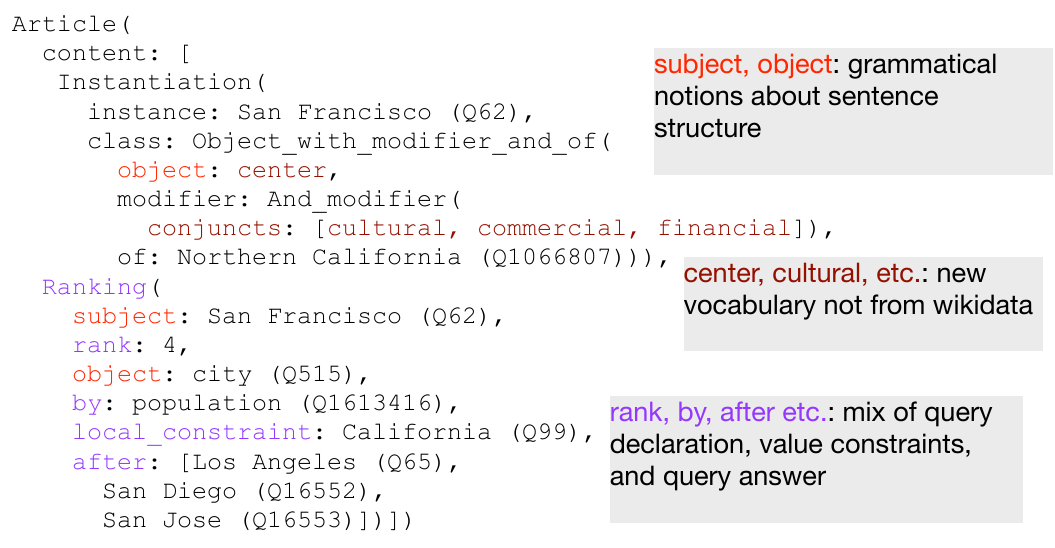}
\caption{Annotated version of the San Francisco example constructor in \cite{Vrandecic20}; see text for discussion.}
\label{fig:SForignal}
\end{figure}

\begin{figure}[h]
\centering
\includegraphics[width = 0.6\textwidth]{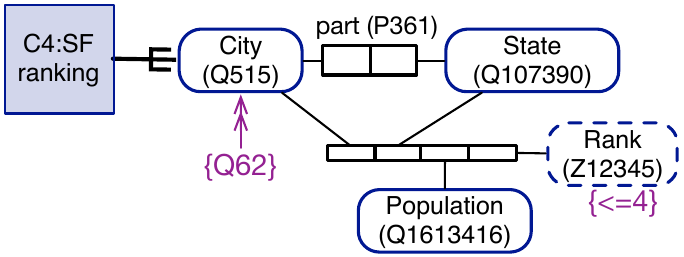}
\caption{Revised San Francisco constructor, diagrammatically rendered in \ourCL~notation.}
\label{fig:SFmodified}
\end{figure}

Let us first reconsider the San Francisco example in \cite{Vrandecic20}, which is depicted on the left-hand side in Fig.~\ref{fig:SForignal}, with colour-coding added for analysis. It exhibits three issues, as indicated on the right-hand side of the figure. First, the pointers to linguistic realisation are supposed to happen later in the pipeline after content selection. Second, the `no new vocabulary' design rule was chosen; hence, those ``{\tt cultural,} ...'' cannot be included as such and must be incorporated in a different way, being by means of Wikidata content. Third, the ranking part must be moved to a function on Wikifunctions that will compute the ranking on the fly. 
Modelling the remainder in \ourCL~is, however, not trivial at present but can be resolved in different ways. First, San Francisco currently has no relation---neither the `{\tt of}' nor another---to {\tt Northern California} (Q1066807) in Wikidata; rather, it is asserted in Wikidata that it is an instance of County of California (Q13212489). Let us liberally use {\sf part of} (P361) and since the ranking is calculated over the state, to relate the city to {\sf State} (Q107390). The adjusted San Frnacisco constructor in \ourCL~diagrammatic notation is shown in Fig.~\ref{fig:SFmodified}.

Let us also revisit Douglas Adams of Example~\ref{ex:DA}, which falls in the same category of information request as the templates for Malala and of Marie Curie from the Scribunto-based proof-of-concept implementation (see fn.~4). That {\tt Constructor-DA} is an instance-level constructor and instead of the perspectivist, object-focussed constructor, the constructor in \ourCL~in text-based format in shorthand notation is as follows, with an optional annotation for illustration:
\begin{verbatim}
 IC:C5(                              // about Writer Douglas Adams
   P106(r1,r2), r1:Q5, r2:Q18844224, 
   P136(r3,r4), r3:Q18844224, r4:24925, 
   Q5={Q42})
   \end{verbatim}
One can generalise from this instance constructor to a constructor type by simply removing the instance selector in the last line, and generalise it further to playwrights of any genre (only the first line), and further generalised to a constructor for humans with occupations by removing the {\tt Q18844224} role filler, or to retrieve all occupations of Douglas Adams by removing the {\tt P136} line and the the {\tt Q18844224} role filler.

\begin{figure}[h]
\centering
\includegraphics[width = 1.0\textwidth]{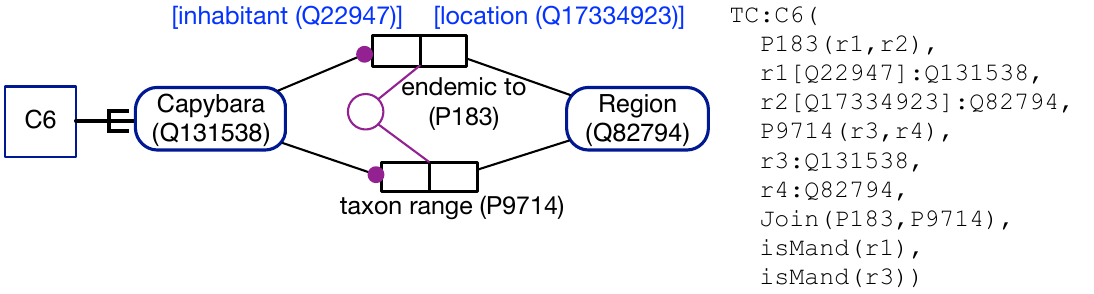}
\caption{Capybara constructor (with placeholder constructor identifier), diagrammatically and textually, demonstrating the two additional features as well as the visualisation of the role names. All Q and P items are valid items in that they already exist in Wikidata.}
\label{fig:capybara}
\end{figure}

Finally, we revisit the capybaras living in a region\footnote{From the sample sentences at \url{https://meta.wikimedia.org/wiki/Abstract_Wikipedia/Sample_sentences}, d.d. 16 Dec 2022.}, for which there are two relevant P-items in Wikidata that we want to 
combine because they are very similar if not synonymous, hence use the 
{\tt Join} option, and with the hard requirement that capybara needs to be in the verbalisation in the target language, which is depicted in Fig.~\ref{fig:capybara}. Thus, it can be represented as per requirements specification.
Additional examples are included in Appendix~\ref{app:ex}, including constructors in Spanish rendering, the back-end natural language-independent rendering with identifiers only, and combining constructors.

\section{Discussion}
\label{sec:disc}

This paper presented a novel modelling language for content selection from databases, be they relational or triple stores, and inclusive of functions, therewith going beyond database view specifications. Yet, while modelling information, it is also distinct from conceptual data modelling languages for it does not strictly require data integrity constraint specification. In addition, one may argue there are additional requirements `for the 21st century', such as the expected use in a setting where there is a stronger demand on multilinguality, openness, collaboration, and sharing. There are obviously also things it cannot do, by design. We will first compare it to related work and then elaborate on the broader setting and for the content selection language.

\subsection{Comparison to related work}

Concerning related work, there is no multilingual modular conceptual modelling language or graphical query builder, as can be seen from comparison included in Table~\ref{tab:modelVis}. 
The three main conceptual data modelling languages included are EER, UML class diagrams, and ORM. We also include three visual SPARQL query builders, since in a way \ourCL~will be capable to function as such as well, albeit also with functions that is not part of SPARQL query building. The three graphical query builders included are VSB\footnote{\url{https://leipert.github.io/vsb/}}, the RDF explorer \cite{Vargas19} that also has a web-based interface\footnote{\url{https://www.rdfexplorer.org/}}, and the early web-based {\sc Wonder} system \cite{CKNRS10} whose graphical rendering was inspired by ORM notation. 

Neither is multilingual nor are they modular. The new online multilingual tool TexToData \cite{Brdjanin23} is multilingual in so far as that it takes text in any natural language, sends it to Yandex machine translation to English, creates the model, and then generates a back translation so that one obtains a draft model in one's input language. Thus, there are multiple models in multiple single languages rather than multilingual modelling where even the elements are multilingual. Regarding modules, while the query builders are not modular by design and it has not been pursued beyond saving queries \cite{CKNRS10}, there are extensions for modularity of conceptual data models. Such modularity is either geared toward scalability of the graphical notations (see \cite{KK21} for a recent overview) or as conceptual model interoperability scenario \cite{Fillottrani11} where the models-to-integrate are recast as modules. Therefore, it received a ``$\pm$'' in the comparison rather than an outright ``--''.  Further, UML class diagrams strictly speaking do not have instances, but the models can link up to object models, and ORM permits derivation rules as a sort of functions and joins over fact types (but not entity types), earning them a ``$\pm$'' each for those features.

\begin{table}[h]
	\centering
	\caption{Comparison of \ourCL~to common conceptual data modelling languages and to a representative selection of graphical SPARQL query builders;  --: no; $\pm$: partial or extensions have been proposed; +: yes; R.E.: RDF Explorer.}			
		\begin{tabular}{|p{3.5cm}|c|c|c||c|c|c||c|}
\hline
{\bf Feature} & {\bf EER} & {\bf UML} & \textbf{ORM 2} & \textbf{WONDER} & \textbf{VSB} & \textbf{R. E.} & \textbf{\ourCL} \\
		\hline \hline
Multilinguality & -- & -- & -- & -- & -- & -- & + \\ \hline
Modularity  & $\pm$ & -- & $\pm$ & -- & -- & -- & + \\ \hline
Class-based & + & + & + & +   & + & + & + \\ \hline
Instances & -- & $\pm$ & -- & -- & -- & + & + \\ \hline
Role naming & + & + & + & -- & -- & -- & + \\ \hline 	
Functions  & -- & + & $\pm$  & -- & -- & -- & + \\ \hline 	
Mandatory and optional & + & + & + & -- & + & -- & + \\ \hline 	
Value constraint & -- & + & + & + & + & + & + \\ \hline 	
Join/merge & --  & -- & $\pm$ & + & + & -- & + \\ \hline 
Disjointness, cardinality, role constraints etc etc. & + & + & + & -- & -- & -- & -- \\ \hline 
`Perspectivist' & + & +  & -- & -- & + & -- & -- \\ \hline 	
Positionalist & +  & +  & + & -- & -- & -- & + \\ \hline 
		\end{tabular}
	\label{tab:modelVis}
\end{table}

It is, of course, possible to arrive at a different abstract representation language, be it based on a different set of requirements or ways to `shape' the elements. The diagrammatic notation was intended to reuse as much as possible an existing notation with open source code (ORM2 \cite{Halpin08} in this case) in order to ease implementation of the graphical notation, but finalising the modifications took multiple iterations. With the logic-based reconstruction, it would also be possible to devise a different notation if users wish to do so. Users also can ignore the graphical notation entirely and focus instead on either of the two textual notations. Also here, it is possible to devise another notation. Either way, any implementation requires rendering all the item identifiers with their respective label in any interface, for it would be very impractical to model with identifiers only. This is an issue to be resolved in the tooling interface.

\subsection{\ourCL~in the context of Abstract Wikipedia and beyond}

The \ourCL~language addresses the declarative part of the content selection specifically and in such a way that it is essentially an enhanced, non-RDF/SQL/ JSON/XML-expert multilingual modular view specification language that is platform-independent. While it indeed couples the elements of \ourCL~to P and Q items, their respective labels easily can reside in a separate array for some tool that does not reside in the Wiki ecosystem. As to the semantics of the language, we could have chosen for an algorithmic approach (analogous to Abstract Wikipedia's template language) by transforming it into SPARQL queries and Zitem function calls, but then \ourCL~would `break' as soon as Wikidata is moved out of Blazegraph and into another technology, or users elsewhere would face the semantic specification hurdle; hence, the logic-based formalisation provided for \ourCL, which can function as a technology-independent pivot.

In addition, and for Abstract Wikipedia and other database specifically, since any NLG technology requires the same sort of access to the source content, instead of requiring each NLG pipeline developer to specify their own constructor language for content selection, this can now be done once fo all such tools. For Abstract Wikipedia specifically, it means that users can write a constructor {\em once}, and thus {\em regardless of} which NLG technique is used to convert it into natural language---be this GF \cite{Ranta23}, Ninai \cite{Morshed23}, SimpleNLG \cite{Gatt09}, under-resourced Niger-Congo B languages \cite{Mahlaza22}, or yet another technique and tool---and {\em regardless of} which data store technology Wikidata is housed in.

Once constructors have been selected and represent content in a systematic and actionable manner, and further down the pipeline, generation and verbalisation considerations need to be addressed.  First, let us consider the semantics.  The semantics, via role assignment, needs to address two types of knowledge: world knowledge and linguistic knowledge, both relevant for realisation.  We foresee two types of roles: ontological roles, for lack of a better term, and linguistic roles (not to be confused with syntactic roles\footnote{By syntactic roles we mean roles such as subject, object, instrumental, etc. which have to do with syntactic aspects and, therefore, language- dependent verbalization.}) \cite{Keet23}.
Both types of roles should be assigned in the early stages of the constructor creation. For ontological roles, this should happen while content selection is effected. We are referring here to any world and ontology knowledge that is relevant for the correct realisation. 
Further down the pipeline, linguistic roles should be assigned.  Semantic roles, in the linguistic sense, are concerned with abstract roles such as agent, patient, material, etc.  These roles are usually attached to arguments and complements and they ensure the correct realisation of the predicate, in the present context, the constructor. At this step one might want to type these roles as mandatory or optional.  In the context of Abstract Wikipedia, the mandatory feature is most likely irrelevant, as the goal is to represent the totality of the content.  But, in other applications it might well be needed.
Syntax comes at the language-dependent realization stage, along with  phonological and/or orthographic adjustments and generation of referring expressions, in steps 5 and 6 in Fig~\ref{fig:pipeline}.
It might also influence ordering of the components to be realised.  Syntactic roles, such as ergative subject, absolutive object, allative argument, etc.  are realised via language-specific rules and features that combine information from semantic roles and features (such as animacy) as well as syntactic restrictions and requirements.

As  mentioned before, linguistic roles will be added after the content selection phase that we are focussing on here with \ourCL. Further down in the pipeline, one may want to add that {\sf r1} is the actor and {\sf r2} the undergoer. This can be added to the constructor or a new one may be created. A possible extension to the constructor could be as follows, if one were to choose to extend the constructor for content selection
to that of linguistic realisation rather than creating a separate constructor and mapping the two: 
\begin{verbatim}
 TC:C1:L( 
   P40(r1,r2), r1:Q7566, r2:Q29514218;
   r1-Actor,
   r2-Undergoer)
\end{verbatim}
Note that the constructor's name is modified, to allow for 1:n between the content constructor and other `constructors'---or: the linguistics-oriented abstract representation---elsewhere in the NLG pipeline. Whether that information is padded into the content constructor like illustrated above, or addressed differently is out of this paper's scope. Likewise out of scope but perhaps useful in visualising the pipeline: after the possible linguistic additions, one can move on to  syntax trees or template specifications. Sample templates for sentences that could be generated from it, where the text in ``[ ]'' is automatically adjusted in the linearisation (and the notation simplified from Abstract Wikipedia's template language for reasons of space and explanation/scope):
\begin{compactitem}
\item The [parent/parents] of \{Q29514218\} [is/are] \{Q7566\}.
\item \{Q7566\} [is/are] the [parent/parents] of {Q29514218}.
\item \{Q7566\} \{P40\} \{Q29514218\}.
\item \{Q29514218\} [inverse \{P40\}] \{Q7566\}.
\end{compactitem}

Last, but not least, one may want to automate generating draft constructors to augment constructor authoring by community members. This could be by, among others, assisting graph navigation in selecting content from Wikidata or trying to induce them from more or less complicated sample sentences. For instance, in a sentence ``The Japanese destroyers Makigumo and Akigumo finally finished off Hornet with 4 24-inch (610 mm) Long Lance torpedoes.'', such a prospective algorithm should recognise the named entities, such as Makigumo, what they instantiate, it needs a function to convert between inches and millimetres, and it needs to check that such entities exist in Wikidata and Wikifunctions. They are tasks that will contribute to the success of \ourCL, as well as depend on \ourCL\ for their successful algorithm design.

\section{Conclusion}
\label{sec:concl}

This paper presented a novel modelling language, \ourCL, with three key novelties 1) multilingual modelling, 2) content selection covering declarative content and functions, and 3) inclusion of both classes and instances. It was developed following a rigorous language design process with stakeholder consultation. 
The preliminary evaluation showed that it met these and other requirements, so that it may be useful both in the early stage of the NLG pipeline of Abstract Wikipedia as well as other contexts. 
Constructors built via \ourCL~facilitate community contributions to a multilingual Wikipedia and it meets the constraints provided by the platform, also enabling people of less resourced and less documented languages to specify and generate content for Wikipedia. 

Current and future work includes soliciting community feedback and a thorough evaluation of \ourCL. It also involves looking ahead to the steps after content selection and linking linguistic and syntactic roles to the {\tt Role} constructor of \ourCL. A tool to assist with authoring constructors is also among the plans.

\subsubsection*{Acknowledgements} We are grateful for for the participants' contribution to the discussions in the various meetings; the meeting participants were: Kutz Arrieta, Maria Keet, James Forrester, Ariel Gutman, Cory Massaro, Arthur Lorenzi, Denny Vrandecic, Ellen Dodge, Nick Wilson. 
This work was financially supported in part by the National Research Foundation (NRF) of South Africa (Grant Number 120852) (MK) and Google.org (KA).

\bibliography{cosmobib}
\bibliographystyle{splncs04}

\appendix

\section{Appendix: Formal Syntax Definition for Longform notation}
\label{app:longsyntax}

The following rules constitute the formal specification of \ourCL\ longform syntax in BNF notation. Terminal symbols are specified in \textbf{bold} for letters, and in \textbf{\underline{underline bold}} for special characters like '':``, '',``, ''(`` and others. The variable nonterminal symbol represents a typical identifier in a programming language, {\em ie} any sequence of letters and numbers starting with a letter. (It has CSMxxx identifiers at its core, but we use the English labels here for readability.)

\begin{align*}
 \langle \text{declarations} \rangle ::=\ &  ( \langle \text{typeConstructor}\rangle \mid \langle \text{instanceConstructor} \rangle \mid \\ &  \langle \text{partOf} \rangle \mid \langle \text{instanceOf}    \rangle \mid \langle \text{subConstructorOf} \rangle ) {^\ast}\\
 \langle \text{typeConstructor}\rangle ::=\ & \textbf{TypeConstructor}\ \mathbf{\underline{:}}\ \langle \text{variable}\rangle \ \mathbf{\underline{(}}\ \langle \text{typeDefinitions} \rangle \ \mathbf{\underline{)}}\\
 \langle \text{typeDefinitions} \rangle ::=\ & (\ \langle \text{predicate}\rangle\ \mathbf{\underline{,}}\ \langle \text{role}\rangle\ \mathbf{\underline{,}}\ \langle \text{role}\rangle\\ & (\mathbf{\underline{,}}\ \langle \text{function} \rangle){^\ast} \ \ (\mathbf{\underline{,}}\ \langle \text{join} \rangle){^\ast} \\
 & (\mathbf{\underline{,}}\ \langle \text{mandatory} \rangle){^\ast} \ \ (\mathbf{\underline{,}}\ \langle \text{instantiation} \rangle) {^\ast} \ ){^+}\\ 
 \langle \text{predicate}\rangle ::=\ & \textbf{Property} \ \mathbf{\underline{(}}\ \langle \text{PItem} \rangle \ \mathbf{\underline{(}}\ \langle \text{variable}\rangle \ \mathbf{\underline{,}}\ \langle \text{variable}\rangle \ \mathbf{\underline{)}}\ \ \mathbf{\underline{)}}\\
 \langle \text{role}\rangle ::=\ & \langle \text{variable} \rangle\ [\ \mathbf{\underline{[}} \langle \text{name}\rangle \mathbf{\underline{]}}\ ]\ \mathbf{\underline{:}}\  
  (\textbf{ObjectType}\mid \textbf{Object})\\ & \mathbf{\underline{(}}\ \langle \text{QItem}\rangle\ \mathbf{\underline{)}}\ [\ \mathbf{\underline{\{}}\ \langle \text{valueConstraint} \rangle\ \mathbf{\underline{\}}} \ ] \\
\langle \text{function}\rangle ::=\ & \textbf{Function}\ \mathbf{\underline{(}}\ \langle \text{ZItem}\rangle\ \mathbf{\underline{(}}\ \langle \text{arguments} \rangle\ \mathbf{\underline{)}}\ \mathbf{\underline{)}}\\
& [\ \mathbf{\underline{\{}}\ \langle \text{valueConstraint} \rangle\ \mathbf{\underline{\}}} \ ] \\  
\langle \text{arguments}\rangle ::=\ & \langle \text{QItem}\rangle\ (\ \mathbf{\underline{,}}\ \langle \text{QItem}\rangle\ ){^\ast}\\
\langle \text{join}\rangle ::=\ & \textbf{Join}\  \mathbf{\underline{(}}\  (\ \langle \text{PItem}\rangle\ \mathbf{\underline{,}}\ \langle \text{PItem}\rangle\ \mid\ 
 \langle \text{QItem}\rangle\ \mathbf{\underline{,}}\ \langle \text{QItem}\rangle\ )\ \mathbf{\underline{)}} \\
 \langle \text{mandatory}\rangle ::=\ & \textbf{IsMandatory}\  \mathbf{\underline{(}}\ \langle \text{variable}\rangle \ \mathbf{\underline{)}}\\ 
\langle \text{instantiation}\rangle ::=\ & \textbf{ObjectType}\  \mathbf{\underline{(}}\  \langle \text{QItem}\rangle\  \mathbf{\underline{)}}\ 
 \mathbf{\underline{=}}\ \langle \text{QItem}\rangle\\
\langle \text{instanceConstructor}\rangle ::=\ & \textbf{InstanceConstructor}\ \mathbf{\underline{:}}\ \langle \text{variable}\rangle \ \mathbf{\underline{(}}\ \langle \text{instDefinitions} \rangle \ \mathbf{\underline{)}}\\
\langle \text{instDefinitions}\rangle ::=\ & (\ \langle \text{predicate}\rangle\ \mathbf{\underline{,}}\ \langle \text{role}\rangle\ \mathbf{\underline{,}}\ \langle \text{role}\rangle\\ & (\mathbf{\underline{,}}\ \langle \text{function} \rangle){^\ast} \ \ (\mathbf{\underline{,}}\ \langle \text{join} \rangle){^\ast} \\
 & (\mathbf{\underline{,}}\ \langle \text{mandatory} \rangle){^\ast} \ \ (\mathbf{\underline{,}}\ \langle \text{instantiation} \rangle){^+} \ ){^+}\\ 
 \langle \text{partOf}\rangle ::=\ & \textbf{PartOf} \ \mathbf{\underline{(}}\ \langle \text{variable}\rangle\ \mathbf{\underline{,}}\  \langle \text{variable}\rangle\ \mathbf{\underline{)}}\\
  \langle \text{instanceOf}\rangle ::=\ & \textbf{InstanceOf} \ \mathbf{\underline{(}}\ \langle \text{variable}\rangle\ \mathbf{\underline{,}}\  \langle \text{variable}\rangle\ \mathbf{\underline{)}}\\
    \langle \text{subConstructor}\rangle ::=\ & \textbf{SubConstructorOf} \ \mathbf{\underline{(}}\ \langle \text{variable}\rangle\ \mathbf{\underline{,}}\  \langle \text{variable}\rangle\ \mathbf{\underline{)}}\\
\end{align*}

\section{Appendix: Shorthand notation}
\label{app:shortnotation}

The longform as compared to a shorthand notation is included in Table~\ref{tab:longshort}. 

\begin{table}[b]
	\centering
	\caption{Longform notation compared to shorthand notation for declaring constructors.}			
		\begin{tabular}{|p{6.8cm}|p{5cm}|}
\hline
 {\bf Longform notation (English)} & {\bf Shorthand notation (English)}  \\
		\hline \hline
 {\tt Object(Q)} or {\tt ObjectType(Q)}& {\tt Q} \\ \hline  	
 {\tt Function(Z(Q1, $\ldots$, Qn))}& {\tt Z(Q1,$\ldots$,Qn)}  \\ \hline  	
 {\tt Property(P(Role1,Role2))} & {\tt P} \\ \hline  
 {\tt TypeConstructor:T()} & {\tt TC} \\ \hline  	
 {\tt InstanceConstructor:O()}&{\tt IC}  \\ \hline  	
{\tt SubConstructorOf(T1,T2)} & {\tt SubTC} \\ \hline  	
 {\tt InstanceOf(O,T)}& {\tt InstOf} \\ \hline  	
{\tt role:Object(Q)} or {\tt role:objectType(Q)} or {\tt Function(Z)} & {\tt r:Q} \\ \hline  
 {\tt PartOf(O1,O2)}& {\tt Po} \\ \hline  	
{\tt TypeConstructor:T({\sl definition})} & {\tt TC:T({\sl definition})} \\ \hline  
 {\tt InstanceConstructor:O({\sl definition})}& {\tt IC:O({\sl definition})} \\ \hline  
{\tt ObjectType\{ValueConstraint\}} or {\tt Object\{ValueConstraint\}}  or {\tt Function\{ValueConstraint\}} & {\tt Q\{\}} or {\tt F\{\}} \\ \hline  
 {\tt Role[Name]:Object(Q)} or {\tt Role[Name]:ObjectType(Q)}& {\tt r[]:Q} \\ \hline  	
{\tt Join(Q1,Q2)} or {\tt Join(P1,P2)}& {\tt Join(Q1,Q2)} or {\tt Join(P1,P2)} \\ \hline  	 		 	
 {\tt IsMandatory(Role)}& {\tt IsMand(r)} \\ \hline  	
{\tt ObjectType(Q1)=\{Q2\}} & {\tt Q1=\{Q2\}} \\ \hline  	  	 	 		 		 		 		 
		\end{tabular}
	\label{tab:longshort}
\end{table}

The formal specification of the shorthand notation is given by the following rules. 

\begin{align*}
 \langle \text{declarations} \rangle ::=\ &  ( \langle \text{typeConstructor}\rangle \mid \langle \text{instanceConstructor} \rangle \mid \\ &  \langle \text{partOf} \rangle \mid \langle \text{instanceOf}    \rangle \mid \langle \text{subConstructorOf} \rangle ){^\ast}\\
 \langle \text{typeConstructor}\rangle ::=\ & \textbf{TC}\ \mathbf{\underline{:}}\ \langle \text{variable}\rangle \ \mathbf{\underline{(}}\ \langle \text{typeDefinitions} \rangle \ \mathbf{\underline{)}}\\
 \langle \text{typeDefinitions} \rangle ::=\ & (\ \langle \text{predicate}\rangle\ \mathbf{\underline{,}}\ \langle \text{role}\rangle\ \mathbf{\underline{,}}\ \langle \text{role}\rangle\\ & (\mathbf{\underline{,}}\ \langle \text{function} \rangle){^\ast} \ \ (\mathbf{\underline{,}}\ \langle \text{join} \rangle){^\ast} \\
 & (\mathbf{\underline{,}}\ \langle \text{mandatory} \rangle){^\ast} \ \ (\mathbf{\underline{,}}\ \langle \text{instantiation} \rangle){^\ast} \ ){^+}\\ 
 \langle \text{predicate}\rangle ::=\ & \langle \text{PItem} \rangle \ \mathbf{\underline{(}}\ \langle \text{variable}\rangle \ \mathbf{\underline{,}}\ \langle \text{variable}\rangle \ \mathbf{\underline{)}}\\
 \langle \text{role}\rangle ::=\ & \langle \text{variable} \rangle\ [\ \mathbf{\underline{[}} \langle \text{name}\rangle \mathbf{\underline{]}}\ ]\ \mathbf{\underline{:}}\  
  \\ & \mathbf{\underline{(}}\ \langle \text{QItem}\rangle\ \mathbf{\underline{)}}\ [\ \mathbf{\underline{\{}}\ \langle \text{valueConstraint} \rangle\ \mathbf{\underline{\}}} \ ] \\
\langle \text{function}\rangle ::=\ & \langle \text{ZItem}\rangle\ \mathbf{\underline{(}}\ \langle \text{arguments} \rangle\ \mathbf{\underline{)}}\\
& [\ \mathbf{\underline{\{}}\ \langle \text{valueConstraint} \rangle\ \mathbf{\underline{\}}} \ ] \\  
\langle \text{arguments}\rangle ::=\ & \langle \text{QItem}\rangle\ (\ \mathbf{\underline{,}}\ \langle \text{QItem}\rangle\ ){^\ast}\\
\langle \text{join}\rangle ::=\ & \textbf{Join}\  \mathbf{\underline{(}}\  (\ \langle \text{PItem}\rangle\ \mathbf{\underline{,}}\ \langle \text{PItem}\rangle\ \mid\ 
 \langle \text{QItem}\rangle\ \mathbf{\underline{,}}\ \langle \text{QItem}\rangle\ )\ \mathbf{\underline{)}} \\
 \langle \text{mandatory}\rangle ::=\ & \textbf{IsMand}\  \mathbf{\underline{(}}\ \langle \text{variable}\rangle \ \mathbf{\underline{)}}\\ 
 \langle \text{instantiation}\rangle ::=\ &  \langle \text{QItem}\rangle\   
\mathbf{\underline{=}}\ \langle \text{QItem}\rangle\\
\langle \text{instanceConstructor}\rangle ::=\ & \textbf{IC}\ \mathbf{\underline{:}}\ \langle \text{variable}\rangle \ \mathbf{\underline{(}}\ \langle \text{instDefinitions} \rangle \ \mathbf{\underline{)}}\\
\langle \text{instDefinitions}\rangle ::=\ & (\ \langle \text{predicate}\rangle\ \mathbf{\underline{,}}\ \langle \text{role}\rangle\ \mathbf{\underline{,}}\ \langle \text{role}\rangle\\ & (\mathbf{\underline{,}}\ \langle \text{function} \rangle){^\ast} \ \ (\mathbf{\underline{,}}\ \langle \text{join} \rangle){^\ast} \\
 & (\mathbf{\underline{,}}\ \langle \text{mandatory} \rangle){^\ast} \ \ (\mathbf{\underline{,}}\ \langle \text{instantiation} \rangle){^+} \ ){^+}\\ 
 \langle \text{partOf}\rangle ::=\ & \textbf{Po} \ \mathbf{\underline{(}}\ \langle \text{variable}\rangle\ \mathbf{\underline{,}}\  \langle \text{variable}\rangle\ \mathbf{\underline{)}}\\
  \langle \text{instanceOf}\rangle ::=\ & \textbf{InstOf} \ \mathbf{\underline{(}}\ \langle \text{variable}\rangle\ \mathbf{\underline{,}}\  \langle \text{variable}\rangle\ \mathbf{\underline{)}}\\
    \langle \text{subConstructor}\rangle ::=\ & \textbf{SubTC} \ \mathbf{\underline{(}}\ \langle \text{variable}\rangle\ \mathbf{\underline{,}}\  \langle \text{variable}\rangle\ \mathbf{\underline{)}}\\
\end{align*}

\clearpage

\section{Appendix: More examples}
\label{app:ex}

The following examples cover: additional natural languages, the underlying pivot with the CSMxxx identifiers, long form vs shorthand notation, and compostionality diagrammatically.

\paragraph{Same constructor rendered in different natural languages}  in accordance with 
 Table~\ref{tab:wikimap}.
\begin{multicols}{2}
\noindent Longform in English
\begin{scriptsize}
\begin{verbatim}
TypeConstructor:C1(
   Property(P40(r1,r2)),     
   r1:ObjectType(Q7566), 
   r2:ObjectType(Q29514218))

InstanceOf(C2, C1)

InstanceConstructor:C2(
   Property(P40(r1,r2)),     
   r1:ObjectType(Q7566), 
   r2:ObjectType(Q29514218), 
   ObjectType(Q29514218)={Q62070381})

SubConstructorOf(C3, C1)
   
TypeConstructor:C3(
   Property(P40(r1,r2)),     
   r1:ObjectType(Q7566), 
   r2:ObjectType(Q29514218),
   Function(Z12345(Q29514218)))
\end{verbatim}   
\end{scriptsize}

\vspace{1ex}
\noindent Longform in Spanish
\begin{scriptsize}
\begin{verbatim}
ConstructorDeTipo:C1(
   Propriedad(P40(r1,r2)),     
   r1:TipoDeEntidad(Q7566), 
   r2:TipoDeEntidad(Q29514218))

InstanciaDe(C2, C1)

ConstructorDeInstancia:C2(
   Propriedad(P40(r1,r2)),     
   r1:TipoDeEntidad(Q7566), 
   r2:TipoDeEntidad(Q29514218), 
   TipoDeEntitdad(Q29514218)={Q62070381})

SubConstructorDe(C3, C1)
   
ConstructorDeTipo:C3(
   Propriedad(P40(r1,r2)),     
   r1:TipoDeEntidad(Q7566), 
   r2:TipoDeEntidad(Q29514218),
   Funcion(Z12345(Q29514218)))
\end{verbatim}   
\end{scriptsize}
\end{multicols}

\paragraph{Same constructor in longform and shorthand notation} in accordance with 
 Table~\ref{tab:longshort}.
\begin{multicols}{2}
\begin{scriptsize}
\begin{verbatim}
InstanceConstructor:C5(                       
   Property(P106(r1,r2)), 
   r1:ObjectType(Q5), 
   r2:ObjectType(Q18844224), 
   Property(P136(r3,r4)), 
   r3:ObjectType(Q18844224), 
   r4:ObjectType(Q24925), 
   ObjectType(Q5)={Q42})

IC:C5(                       
   P106(r1,r2), 
   r1:Q5, 
   r2:Q18844224, 
   P136(r3,r4), 
   r3:Q18844224, 
   r4:Q24925, 
   Q5={Q42})
\end{verbatim}   
\end{scriptsize}
\end{multicols}

\clearpage

\paragraph{Constructor and sub-constructor in longform rendering and its behind-the-scenes \ourCL\ encoding as pivot} in accordance with Table~\ref{tab:wikimap}. The pivot with the CSM numbers are not intended for human consumption, only back-end computational processing.
\begin{multicols}{2}
\begin{scriptsize}
\begin{verbatim}
InstanceConstructor:C5(                       
   Property(P106(r1,r2)), 
   r1:ObjectType(Q5), 
   r2:ObjectType(Q18844224), 
   Property(P136(r3,r4)), 
   r3:ObjectType(Q18844224), 
   r4:ObjectType(Q24925), 
   ObjectType(Q5)={Q42})
  
SubConstructorOf(C3, C1)  

CSM007:C5(                       
   CSM003(P106(r1,r2)), 
   r1:CSM002(Q5), 
   r2:CSM002(Q18844224), 
   CSM003(P136(r3,r4)), 
   r3:Q18844224, 
   r4:Q24925, 
   CSM002(Q5)={Q42})
   
CSM008(C3, C1)
\end{verbatim}   
\end{scriptsize}
\end{multicols}

\paragraph{Constructor with value constraints} in textual notation and where all identifiers have been resolved to the labels in the selected language.  

\begin{scriptsize}
\begin{verbatim}
TypeConstructor:C6(                       
   Property(works(r1,r2)), 
   r1:ObjectType(Student{has scholarship}), 
   r2:ObjectType(University), 
   Property(citizen of(r3,r4)),
   r3:ObjectType(Student),
   r4:ObjectType(Country{Argentina}),
   Function(top three(Student, gpa{>95%}))
)
\end{verbatim}   
\end{scriptsize}

\paragraph{Combining constructors} into a larger, and thus more complex, whole, in diagrammatic notation.  

\begin{figure}[h]
\centering
\includegraphics[width=0.7\textwidth]{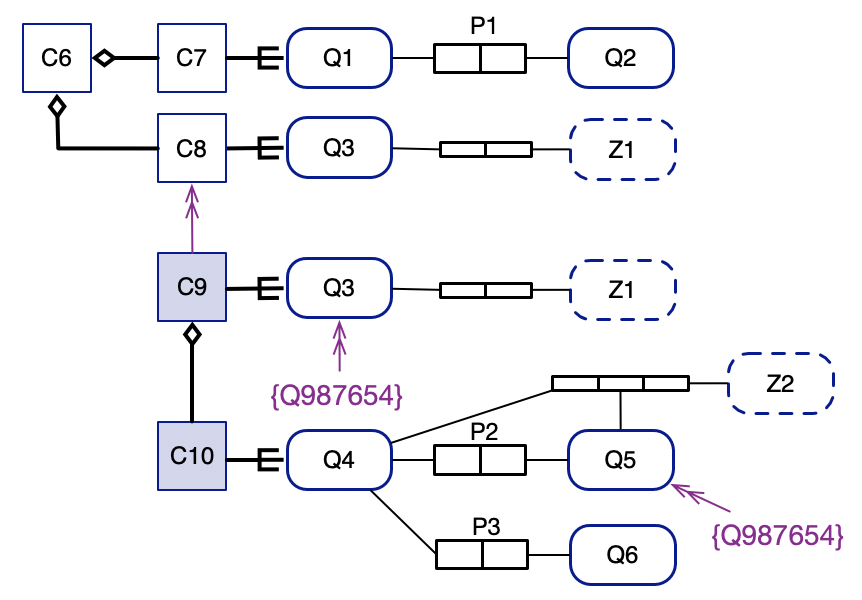}
\caption{A schematic example with various options.}
\label{fig:compose}
\end{figure}

\end{document}